\documentclass{article}

\usepackage{arxiv}

\usepackage[utf8]{inputenc} 
\usepackage[T1]{fontenc}    
\usepackage{hyperref}       
\usepackage{url}            
\usepackage{booktabs}       
\usepackage{amsfonts}       
\usepackage{nicefrac}       
\usepackage{microtype}      
\usepackage{graphicx}
\usepackage{natbib}
\usepackage{doi}
\usepackage{amsmath}
\usepackage{subcaption}
\usepackage{graphicx}
\usepackage{multirow}
\usepackage{amssymb}
\usepackage[title]{appendix}

\title{Prediction of liquid fuel properties using machine learning models with Gaussian processes and probabilistic conditional generative learning}

\author{ \href{https://orcid.org/0000-0001-6036-8534}{\includegraphics[scale=0.06]{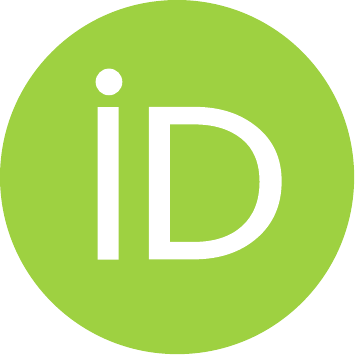}\hspace{1mm}Rodolfo S. M. Freitas} \\
	Dept. Mechanical Engineering\\
	COPPE/ Federal University of Rio de Janeiro\\
	RJ 21941-598, Rio de Janeiro, Brazil\\
	\texttt{rodolfosmfreitas@gmail.com} \\
	\And
	\href{https://orcid.org/0000-0002-2155-6185}{\includegraphics[scale=0.06]{orcid.pdf}\hspace{1mm}{\'A}gatha P. F. Lima} \\
	Dept. Mechanical Engineering\\
	COPPE/ Federal University of Rio de Janeiro\\
	RJ 21941-598, Rio de Janeiro, Brazil\\
	\texttt{agatha.pires@coppe.ufrj.br} \\
	\And
	\href{https://orcid.org/0000-0001-7292-9490}{\includegraphics[scale=0.06]{orcid.pdf}\hspace{1mm}Cheng Chen} \\
	School of Engineering and Materials Science\\
	Queen Mary University of London\\
	Mile End Road, London E1 4NS, UK\\
	\texttt{cheng.chen@qmul.ac.uk} \\
	\And
	\href{https://orcid.org/0000-0001-8035-9651}{\includegraphics[scale=0.06]{orcid.pdf}\hspace{1mm}Fernando A. Rochinha} \\
	Dept. Mechanical Engineering\\
	COPPE/ Federal University of Rio de Janeiro\\
	RJ 21941-598, Rio de Janeiro, Brazil\\
	\texttt{faro@mecanica.coppe.ufrj.br} \\
	\And
	\href{https://orcid.org/0000-0001-9901-7942}{\includegraphics[scale=0.06]{orcid.pdf}\hspace{1mm}Daniel Mira} \\
	Barcelona Supercomputing Center\\
	Barcelona, Spain\\
	\texttt{daniel.mira@bsc.es} \\
	\And
	\href{https://orcid.org/0000-0003-2408-8812}{\includegraphics[scale=0.06]{orcid.pdf}\hspace{1mm}Xi Jiang} \\
	School of Engineering and Materials Science\\
	Queen Mary University of London\\
	Mile End Road, London E1 4NS, UK\\
	\texttt{xi.jiang@qmul.ac.uk} \\
	
	\thanks{The research leading to these results had received funding from the Brazilian National Agency for Petroleum, Natural Gas and Biofuels (ANP) through Programa de Recursos Humanos (PRH) under the PRH 8 - Mechanical Engineering for the Efficient Use of Biofuels, grant agreement numbers F0A5.EDDE.B5C0.3BCB and 2B61.4F5C.A83B.A713.}
}

\date{}

\renewcommand{\shorttitle}{R. S. M. Freitas, {\'A}. P. F. Lima, C. Chen, F. A. Rochinha, D. Mira \& X. Jiang}

\hypersetup{
pdftitle={A template for the arxiv style},
pdfsubject={q-bio.NC, q-bio.QM},
pdfauthor={David S.~Hippocampus, Elias D.~Striatum},
pdfkeywords={First keyword, Second keyword, More},
}

\begin{document}
\maketitle

\begin{abstract}
Accurate determination of fuel properties of complex mixtures over a wide range of pressure and temperature conditions is essential to utilizing alternative fuels. The present work aims to construct cheap-to-compute machine learning (ML) models to act as closure equations for predicting the physical properties of alternative fuels. Those models can be trained using the database from MD simulations and/or experimental measurements in a data-fusion-fidelity approach. Here, Gaussian Process (GP) and probabilistic generative models are adopted. GP is a popular non-parametric Bayesian approach to build surrogate models mainly due to its capacity to handle the aleatory and epistemic uncertainties. Generative models have shown the ability of deep neural networks employed with the same intent. In this work, ML analysis is focused on a particular property, the fuel density, but it can also be extended to other physicochemical properties. This study explores the versatility of the ML models to handle multi-fidelity data. The results show that ML models can predict accurately the fuel properties of a wide range of pressure and temperature conditions.
\end{abstract}


\section{Introduction}

Fossil fuels have been playing a major role in energy supply and liquid fossil fuels have dominated the energy use in transport, which will continue to be so for many decades to come, especially for sectors that are difficult to decarbonise. With the pressing needs of decarbonisation and sustainable energy utilisation, renewable fuels and biofuels are becoming increasingly important \cite{Omari2017,Pelerin2020}. For instance, synthetic fuels like Oxymethylene Dimethyl Ethers (OMEs) have shown
high potential for low-carbon
transport applications due to their capacity to
avoid soot formation~\cite{Pastor2020}. However, the physicochemical properties of these fuels must be known for their rapid integration into current infrastructures for storage, transport and direct injection in combustion engines. This represents a significant challenge, due to the fact that practical fuels are often composed by complex mixtures and vary widely in their chemical compositions depending on the production source and process~\cite{Omari2017}. For example, petroleum diesel is a complex mixture involving molecules with carbon chains that typically contain between 9 and 25 carbon atoms per molecule. To simplify the complex chemical compositions of these fuels, surrogate models have been used to represent the chemical composition and combustion characteristics in practical applications~\cite{PITZ2011330,LAI20111}. In addition, modern combustion engines have to operate at high pressure conditions in order to improve the energy conversion efficiency. Fuel properties at extreme conditions such as high pressure and high temperature conditions, are very difficult to measure and predict~\cite{Pastor2020}, leading to an additional challenge.

Accurate determination of fuel properties of complex mixtures over a wide range of pressure and temperature conditions is essential to adapt the system operation to alternative fuels. In recent years, molecular dynamics (MD) simulations have been used to predict the physicochemical properties of practical fuels including transport properties at supercritical conditions \cite{CHEN201948}. By using equilibrium molecular dynamics (EMD) and nonequilibrium molecular dynamics (NEMD), Yang et al \cite{yang2021molecular,yang2021comparison} predicted the viscosity and thermal conductivity of alkanes (n-decane, n-undecane and n-dodecane). Kondratyuk et al \cite{kondratyuk2020probing,kondratyuk2020transport,kondratyuk2019calculation} performed a serial of MD simulation to study the viscosity of hydrocarbons (1-methylnaphthalene, methylcyclohexane and 2,2,4-trimethylhexane) in high pressure conditions up to 1000 MPa. Caleman et al \cite{caleman2012force} tested the capacity of existing force fields on prediction of properties (density, enthalpy of vaporization, surface tension and heat capacity etc) of organic liquids. Although MD simulations provide molecular details that can be potentially used to accurately predict fuel properties, they are generally expensive in terms of computational costs. In addition, MD predictions also need to be validated against experimental measurements, which can be even more costly especially at extreme conditions. Accordingly, it is not feasible to establish complete and detailed fuel property databases consisting of a wide range of pressure and temperature conditions using either MD simulations or experiments.

Machine learning has great potentials to discover the relation between inputs and outputs in a thermodynamic system directly from the data of complex systems~\cite{FREITAS2020115949}. ML can be a powerful tool to predict fuel properties from chemical compositions of the fuel mixture and/or chemical structures of the fuel molecules. Several works have been devoted to designing ML models capable of predicting complex fuels properties from experimental data. In this regard, ML models obtained accurate predictions of cetane number (CN) compared to experimental data \cite{RAMADHAS20062524, PILOTORODRIGUEZ2013255, MIRABOUTALEBI2016143}. A satisfactory ML approach for modeling the CN of biodiesel based on four operating conditions given by iodine volume (IV), carbon number, double bounds, and saponification value was proposed \cite{NAJAFI12960}. Moreover, ML models were tuned with evolutionary algorithms to predict the CN of biodiesel as a function of its fatty acid methyl ester (FAME) profile {\cite{MOSTAFAEI2018665, BEMANI2020924}}. The predictability of the ML approaches also can be improved by using different optimization algorithms such as teaching-learning based optimization (TLBO), backpropagation, Quasi-Newton and particle swarm optimization (PSO) \cite{BAGHBAN2018620, NOUSHABADI2020465, SANCHEZBORROTO2014877}. Also, ML models have been used for modeling the kinematic viscosity of diesel-derived fuels as a function of their FAMEs profiles \cite{ZHAO1641575, ALVISO2020116844, MENG2014133}. In the last years, Multilayer Perceptron Neural Networks (MLPNNs) have been successfully built to estimate the physicochemical characteristics of biodiesel \cite{Cheenkachorn2006PredictingPO, SOLOMON2015, ROCABRUNOVALDES20159, OLIVEIRA00282} combining different parameters of model inputs. Furthermore, ML models based on state variables such as temperature and pressure showed high potential to obtain physicochemical properties of biodiesel/diesel fuels more accurately~\cite{ZHOU2018, Erylmaz2015PredictionOK, TANZER2410}. Recently, a ML approach based on support vector regression (SVR) was proposed by \cite{LIU2019116091} for predicting the PVT properties of pure fluids (H$_2$O, CO$_2$, and H$_2$) and their mixtures, where the training database is provided by the National Institute of Standards and Technology (NIST) and MD simulations. Also, an ML approach was proposed to assess the macroscopic ECN Spray-A characteristics and predictions of fluid properties for the thermodynamic states found in such conditions \cite{ML_sprayA}. Yet, from our knowledge, little work has been dedicated towards exploring the thermodynamic properties of practical fuels combining MD simulations and ML models. ML can be a powerful tool to predict fundamental fuel properties directly from the chemical compositions of the fuel mixture by using databases from MD simulations or available experimental measurements.


The aim of the study was to demonstrate and validate a ML-MD methodology to predict fundamental properties of liquid fuels. In this approach, the ML models are built from data provided by MD simulations, while a combination of MD and NIST data is used for model assessment and validation. This study is the first attempt of using ML models with Gaussian process regression \cite{GP_RASMUSSEM} and probabilistic conditional generative learning \cite{YANG2019136,PERDIKARIS2019} for the property predictions of single-compound fuels. The ML analysis is focused on fuel density in this study as one of fundamental properties of liquid fuels, though it can easily be extended to other physicochemical properties of relevance for practical applications like viscosity, conductivity or surface tension.

The rest of the paper is organized as follows. Section 2 presents the ML models and the molecular dynamics simulation methodology. Section 3 describes the ML results for typical fuel surrogates of diesel. Finally, Section 4 concludes the study with recommendation for further investigations.

\section{Methodology: Building Machine Learning Models to Describe Physicochemial Properties}

{In order to reduce energy consumption and pollutant formation, supercritical combustion has been increasingly explored in the context of high pressure internal combustion engines and rocket engines \cite{SAFAROV2018870}. Specifically, in supercritical conditions, the devices operate with pressures and temperatures higher than the critical values, which implies that  physicochemical properties of  fluids are quite different from those at liquid conditions \cite{PioroMokryDraper+2011+191+214}.} In such scenarios, the design of devices become more complex, specially due to limitations of replicating flow and combustion in controlled laboratory environments. In order to cope with these challenges, computational models can provide adequate tools for obtaining more accurate predictions of state variables and increase cycle performance in transcritical conditions.

{From a computational fluid dynamics (CFD) perspective, combustion models are built upon the combination of solid and reliable physico/chemical principles with closure models, typically describing physicochemical properties of the fuels and their mixtures using approaches that normally entail uncertainties. {The use of numerical simulations for practical applications encompass a wide range of conditions, resulting in different fundamental problems depending on the nozzle geometry, engine architecture or thermodynamic conditions. A good example is the database from the Engine Combustion Network~\cite{ENC_sprayA} for which different sprays for diesel- and gasoline-like conditions are investigated.} For instance, pressure can go from sub-atmospheric to 2,000 bar, and temperatures from cold to highly preheated conditions. In that context, having accurate values for macroscopic fuel characteristics and properties over such wide variety of spatial and time scales is one of the main challenges for physically-driven methods. That is particularly more dramatic for modern compounds depicting complex chemical compositions, and simplified surrogate fuels \cite{surrogate_Cao} are employed to estimate the properties of the original compounds. That allows the systematic use of controlled experiments and, also, Molecular Dynamics simulations \cite{RAZI2018125,XING202136}. Indeed, here our focus lies on using ML models to leverage such type of simulations when obtaining liquid fuel physicochemical properties. Those properties are generally expressed as functions of local thermodynamic conditions like pressure and temperature, which motivate to refer to closure models such as the Equations of State (EoS). In general, the EoS is embedded in complex CFD simulations resulting in divergence or numerical oscillations when used with traditional methods based on tabular and interpolation schemes \cite{KOUKOUVINIS2020117871}. It is worth to remark that we are seeking for models capable of describing physicochemical properties over a wide range of flow conditions and we expected to observe abrupt changes around critical conditions.}

{Here, we build two different ML models, namely   Gaussian Processes (GP) \cite{GP_RASMUSSEM} and a probabilistic conditional generative approach \cite{PERDIKARIS2019}. We train both in a supervised learning fashion using data produced with expensive MD simulations. Therefore, we rely on their ability to learn from a small amount of data and their capacity of extrapolation. Moreover, we also want to take into consideration the unavoidable uncertainties arising from limited information (epistemic) and from noisy data (aleatoric).}

{GPs have become popular due to its success on being a proxy for  high-fidelity models in different applications \cite{GP_RASMUSSEM,RAISSI0751, SU201797, CHEN200759, YUAN200847, GUERRA2018}. Another well proved ML approach are the so called generative models that explore existing low-dimensional structures capable of explaining high-dimensional data introducing probabilistic latent variables.}

In the remainder of this chapter, we present a brief description of both ML models for a generic property $\gamma (p,T)$ function of pressure and temperature, along with the corresponding training algorithms. For the  training of the models,  we assume the availability of, potentially expensive, dataset comprising input/output pairs $ \{(p,T)_i,\mathbf{\gamma}_i\ \ \ i=1,...,n\}$  generated by an implicit mapping $g$ characterizing the macroscopic thermodynamic relation between the property and the state variables:
\begin{equation}
    {\gamma}= g({p,T};\boldsymbol{\xi}).
    \label{eq:GP}
\end{equation}

\noindent The role of $g$ here is played by upscaling MD simulations or, to a less extent, by  experimental available data. The vector $\boldsymbol \xi$ denotes potential noisy and is often considered a random. In order to keep a compact notation, we refer to the above dataset as $\mathcal{D} = (\mathbf{x},\mathbf{y})$, with $\mathbf{x} \in \mathbb{R}^{2n}$ and $\mathbf{y}\in \mathbb{R}^n$ vectors containing  inputs and outputs. We intentionally do not use the word surrogate to designate any of the two ML models to avoid misleadings. In the combustion technical literature, it is employed to refer to compounds with simpler compositions to replace complex fuels in experimental or numerical analysis.

\subsection{Gaussian process regression}

A GP is an infinite collection of random variables, in which  any finite number of such variables depict a joint Gaussian distribution \cite{GP_RASMUSSEM}.  In line with Bayesian estimation, to approximate $g$ we assign a GP zero mean prior $f(\mathbf{x})$ , i.e., $f \sim GP(f|\mathbf{0},k(\mathbf{x},\mathbf{x}^{'};\boldsymbol{\theta}))$, where $k$ is a kernel  parametrized by a vector of hyper-parameters $\boldsymbol{\theta}$ to be learned from $\mathcal{D}$ and engenders a symmetric positive-definite $n \times n$ covariance matrix $K_{ij}=k(x_i,x_j;\boldsymbol{\theta})$. Instead of choosing the squared exponential form of the kernel as usual \cite{GP_RASMUSSEM}, here, we test some forms of covariance matrix belonging to the Matern family. More specifically, we employ the Mayern 3/2 covariance matrix given as  
\begin{equation}
    k(\mathbf{r}) = \sigma^2 \left(1+ \sqrt{6}\frac{|r|}{l}\right) \text{exp}\left(-\sqrt{6}\frac{|r|}{l}\right)
    \label{eq:cov_matrix}
\end{equation}
\noindent with $\mathbf{r}=\mathbf{x}-\mathbf{x}^{'}$ denoting the distance between different inputs. The hyper-parameters are the standard deviation $\sigma$, and the correlation lengths $\mathbf{l}=\{l_1,l_2,\dots,l_{n_k} \}$, and $n_k$ denotes the dimension of input $\mathbf{r}$. Hence, the hyper-parameters vector reduces to $\boldsymbol{\theta}=\{\mathbf{l},\boldsymbol{\sigma} \}$.

{ We do not follow a fully Bayesian  approach, and obtain the vector of hyper-parameters $\boldsymbol{\theta}$  by maximizing the marginal log-likelihood of the model, i.e.}
\begin{equation}
    \text{log} p(\gamma|\mathbf{x}, \boldsymbol{\theta}) = - \frac{1}{2} \text{log}|\mathbf{K}|-\frac{1}{2}\gamma^T\mathbf{K}^{-1}\gamma-\frac{n}{2}\text{log} 2\pi.
    \label{eq:log_likelihood}
\end{equation}

\noindent  using a conjugate gradient descend method.

The final goal of the  regression is obtaining a predictive model for $\gamma$, which means to compute its value for an untested state $\mathbf{x}_*$ \cite{RAISSI0751}

\begin{equation}
    \mu_*(\mathbf{x}_*)=k_{*n}\mathbf{K}^{-1}\mathbf{y}
\end{equation}
{and}
\begin{equation}
    \sigma_*^2(\mathbf{x}_*)=k_{**}-k_{*n}\mathbf{K}^{-1}k_{*n}^T
\end{equation}

\noindent {where $k_{*n}=[k(\mathbf{x}_*,\mathbf{x}_1),\dots,k(\mathbf{x}_*,\mathbf{x}_n)]$ and $k_{**}=k(\mathbf{x}_*,\mathbf{x}_*)$. The predictions are computed using the posterior mean $\mu_*$, and the uncertainty associated with that predictions is quantified through the posterior variance $\sigma_*^2$. It is worth to mention that in absence of noisy in the training data, the later represents epistemic uncertainty due to lack of data.}

\subsection{Probabilistic conditional generative model}

Now, we explore a  probabilistic conditional generative approach \cite{YANG2019136,PERDIKARIS2019}, that integrates  variational auto-encoders (VAE) \cite{kingma2014autoencoding} and generative adversarial networks (GANs) \cite{goodfellow2014generative}. Moreover, it  employs a probabilistic perspective that enables to take into consideration noisy and limited data from the beginning. It is also capable of dealing with high-dimensionality of inputs and outputs, what is not explored here due to the specific aspects of our needs.

The final goal is to build probabilistic neural networks that follow a conditional probability density function $p(\gamma|(p,T), \mathcal{D})$ learnt from the data. So, the surrogate model can deploy accurate values for the property $\gamma$
by estimating the expectation $\mathbb{E}( \gamma|(p,T),\mathcal{D})$, and also, to quantify the uncertainty associated with that prediction in CFD calculations.

The main ingredient for this approach is the introduction of a vector of latent random variables  aiming at  seeking for a hidden low dimensional structure for explaining the  data structure. In a formal abstract perspective, such latent variables allow us to express the conditional probability associate to the data $\mathcal{D}$, not included in the expression to keep the notion clear, $p(\gamma |p, T)$, as an infinite mixture model through 

\begin{equation}
    p(\gamma |p,T)=\int p(\gamma,\mathbf{z}|p,T) \ d\mathbf{z}=\int p(\gamma|p,T,\mathbf{z})\ p(\mathbf{z}|p,T)\ d\mathbf{z}
    \label{eq:infinite_mixture}
\end{equation}

\noindent where $p(\mathbf{z}|p,T)$ is a prior distribution on the latent variables. The above hierarchical mathematical ansatz, despite being very elegant and rigorous, has to be approximated \cite{PERDIKARIS2019}, where a
regularized adversarial inference framework is proposed and detailed. The final result is a generator model $\gamma=f_{\phi}(p,T,\mathbf{z})$ parametrized by vector $\phi$, like trained deep neural networks. In conjunction with $p(\mathbf{z})$, the statistics of $\gamma$ can be characterized. More specifically, we can  compute its low order statistics via Monte Carlo sampling. It is important to remark that the predictions with the identified probabilistic generator, that, in present context, plays the role of a proxy for  obtaining macroscopic thermodynamic properties of mixtures for pressures and temperatures not contained in $\mathcal{D}$,  is negligible when compared to MD simulations. The mean and variance of the predictive distribution at a new  point $(p^* , T^*)$ are computed as
\begin{equation}
    \mu_{\gamma} (p^* , T^*)=\mathbb{E} [\gamma| p^* , T^*\mathbf{z}]\approx \frac{1}{N_s}\sum_{i=1}^{N_s}\left[f_{\phi}(p^* , T^*,\mathbf{z}_i)\right]
\end{equation}
\begin{equation}
    \sigma_{\gamma}^2(p^* , T^*)=\mathbb{V}\text{ar}[\gamma |p^* , T^*,\mathbf{z}]\approx \frac{1}{N_s}\sum_{i=1}^{N_s}\left[f_{\phi}(p^* , T^*,\mathbf{z}_i) - \mu_{\mathbf{y}}(p^* , T^*)\right]^2,
\end{equation}
{where $\mathbf{z}_i\sim p(\mathbf{z})$, $i=1,\dots,N_s$, and $N_s$ corresponds to the total number of  samples.}

At this point, it is important to clarify that  the predictive uncertainty encoded in $\mathbf{z}$ is due to  noise in the Molecular Dynamics computations originated by numerical approximations and to the potential small amount of data employed in the training process. Therefore, it encapsulates aleatoric and epistemic uncertainties. 

Later, we explore the versatility of the probabilistic ML model  employing the fusion of data produced by MD with experimental data obtained for supercritical behavior of the mixture.

\subsection{Density prediction in EMD simulation}
In this study, all MD simulations are performed in Gromacs package \cite{van2005gromacs} with Transferable Potentials for Phase Equilibria (TraPPE) force field \cite{martin1998transferable}. United-atom molecular description is used in order to reduce the computational cost. Before simulation, 500 molecules are distributed in a box with relatively large edge length of 14 nm to avoid atom's overlap. After energy minimisation, a 500 ps simulation is followed in  isobaric-isothermal ensemble. The temperature is controlled by velocity rescale, and the pressure is maintained by using the Parrinello-Rahman approach \cite{parrinello1981polymorphic}. The time step is set to be 1 fs. The fixed bond length in TraPPE force field is achieved by using LINCS algorithm \cite{hess1997lincs}. The density is averaged over the last 300 ps simulation.

\section{Results and discussion}

{Here, we   demonstrate  the performance of the proposed {methodology.} Despite alternative fuels can be very complex mixtures consisting of hundreds of compounds, we consider  single-component alkanes C$_n$H$_{2n+2}$, so reliable data for model assessment and validation can be used. In general, realistic fuels are usually described by  surrogate models~\cite{CHEN201948} because of  availability of validated chemical mechanisms and experimental measurements. The data to train our ML models consist of  properties of a family of alkanes, ranging from normal to supercritical conditions. More specifically, we construct ML models to characterize density dependency on some operational conditions in which data is not available. As mentioned before, in order to take into consideration unavoidable uncertainties, we approximate the conditional probability $p(\gamma| \mathbf{x}, \boldsymbol{\theta})$, with $\mathbf{x}$ being the input vector with components pressure $p$, temperature $T$ and chemical composition. Moreover, it is worth mentioning here that we consider as the input that characterizes the chemical compositions the number of atoms of carbon $C$ in the fuel molecule, a categorical variable. Also, for the GP learning model, the hyper-parameters vector reduces to $\boldsymbol{\theta}=\{\mathbf{l},\boldsymbol{\sigma} \}$, and for the generative model $\boldsymbol{\theta}$ represents a set of latent variables $z\sim p(z)$}. Here, we employ a one-dimensional latent space with a standard normal prior, $p(z) \sim \mathcal{N}(0, 1)$.

{The fuels considered here correspond to n-octane, n-nonane, n-decane, n-dodecane, and n-hexadecane, operating within conditions ranging from high-pressure nozzle condition to supercritical chamber environment. The dataset used to build the ML models consists of 1200 density values. Specifically, the dataset is composed of 240  densities values of each diesel surrogate fuel at a regular temperature grid $T\in[320, 900]$ K, varying by 20K, and at specific  pressures: $p=[3, 4, 6, 8,10,20,100,150]$ MPa. Also, it is worth remarking that in this dataset we included density values for supercritical regions, more specifically values above the critical temperature ($T_c$) of the compositions, being the critical values for n-octane ($T_c = 569.32 K$), n-nonane ($T_c = 594.55 K$), n-decane ($T_c = 617.7 K$), n-dodecane ($T_c = 658.1 K$), and n-hexadecane ($T_c = 722 K$), which replicate engine-like conditions}. 

{In the learning process, 80\% of the data points are selected randomly to training the ML models. The remaining 20\% are used to testing them. Moreover, the training data set is organized in three subsets with 10\%, 50\%, and 100\% of data available to train the models. The aim here is to evaluate the convergence and impacts of constructing the ML models in a small data regime. Accuracy is measured using the distance between the expected values predicted with the ML models and the predictions computed with the MD simulations.  We check this accuracy computing the $L_2$ mean relative error ($L_{2-MRE}$)}
\begin{equation}
    L_{2-MRE} = \frac{1}{N}\sum_{i=1}^{N}\left(\frac{\rho_i - \hat{\rho}_i}{\rho_i}\right)^2
\end{equation}
{where $\rho_i$ is the density computed with MD simulations, $\hat{\rho}_i$ is the expected ML output and $N$ is the number of test samples. Also, we compute the coefficient of determination ($R^2$-scire) metric \cite{Sanfordbook}}
\begin{equation}
    R^2 = 1 - \frac{\sum_{i=1}^{N} \parallel \rho_i - \hat{\rho}_i \parallel_{2}^{2}}{\sum_{i=1}^{N} \parallel \rho_i - \overline{\rho} \parallel_{2}^{2}}
\end{equation}
{where $\overline{\rho}=\frac{1}{N}\sum_{i=1}^{N}\rho_i$ is the mean density of test samples. The $R^2$-score metric represents the normalized error, allowing the comparison between ML models trained by different data sets, with values close to 1 corresponding to the ML models best accuracy.}

{We obtain the GP regression model of Eq. \eqref{eq:GP} via maximizing the marginal log-likelihood of Eq. \eqref{eq:log_likelihood} using the Mattern 3/2 kernel function, as that shown in Eq. \eqref{eq:cov_matrix}. Also, we have used the gradient descend optimizer L-BFGS \cite{BFGS_LIU} using randomized restarts to ensure convergence to a global optimum. The GP learning model was implemented in GPy: Gaussian Process (GP) framework written in python \cite{gpy2014}.}

{On the other hand, to construct the generative learning model, we departed from  the architecture proposed and validated by Yang and Perdikaris \cite{PERDIKARIS2019}. More specifically, the conditional generative model is constructed using fully connected feed-forward architectures for the encoder and generator networks with 4 hidden layers and 100 neurons per layer, while the discriminator architecture has 2 hidden layers with 100 neurons per layer. The neural networks are constructed by combining try-and-error and Hyperopt algorithm \cite{hyperopt} to search for the hyperparameters that give the lowest $L_{2-MRE}$. All activation uses a hyperbolic tangent non-linearity. The models are trained for 50,000 stochastic gradient descent steps using the Adam optimizer \cite{kingma2017adam} with a learning rate of $10^{-4}$, while fixing a two-to-one ratio for the discriminator versus generator updates. Furthermore, we have also fixed the entropy regularization parameter to $\lambda=1.5$ and $\beta=0.5$, respectively. The proposed model was implemented in TensorFlow v2.1.0 \cite{tensorflow2015-whitepaper}, and computations were performed in single precision arithmetic on a single NVIDIA GeForce RTX 2060 GPU card.}

We also explore some alternatives versions of the above described ML models by proposing fusion with experimental data and the use of multi-fidelity formulations.

\subsection{ML results for typical fuel surrogates}

{Tables \ref{tab:gp_accuracy} and \ref{tab:gans_accuracy} show the coefficient of determination (R$^2$-score) and $L_2$ mean relative error, respectively, for GP and the probabilistic conditional generative models. The accuracy metrics are computed with the test samples. We observe that they are not satisfactory in the small training data scenario, with 10\% of training data.  R2-scores for the GP and conditional generative models in this specific training scenario are 0.8538 and 0.9359, respectively. For a data richer situation, with 50\% of training data, we observe that the  models return good predictions with R$^2$-score higher than $0.99$. Also, we observe that the conditional generative model returns better predictions than the GP model in a small data scenario, with an accuracy of $L_{2-MRE}=2.8989 \times 10^{-3} $ while the GP accuracy is $L_{2-MRE}= 4.7428 \times 10^{-2}$. Finally, with 100\% of the training data, we can see that the surrogate models return excellent predictions with R$^2$-score very near $1.0$ and mean relative errors lower than 0.03\%.}

\begin{table}
\caption{Gaussian Process training accuracy.}
\centering
\begin{tabular}{ccc}
\toprule
\textbf{Train data} & \textbf{$L_{2-MRE}$} & \textbf{R$^2$-score} \\
\midrule
10 \% & $6.2805 \times 10^{-2}$ & $0.8538$  \\
50 \% & $4.7438 \times 10^{-2}$ & $0.9976$   \\
100 \% &	$2.7272 \times 10^{-2}$ & $0.9991$ \\
\bottomrule
\end{tabular}
\label{tab:gp_accuracy}
\end{table}
    
\begin{table}
\caption{Generative model training accuracy.}
\centering
\begin{tabular}{ccc}
\toprule
\textbf{Train data} & \textbf{$L_{2-MRE}$} & \textbf{R$^2$-score} \\
\midrule
10 \% & $4.9316 \times 10^{-2}$ & $0.9359$  \\
50 \% & $2.8989 \times 10^{-3}$ & $0.9983$   \\
100 \% & $2.1409 \times 10^{-3}$ & $0.9990$ \\
\bottomrule
\end{tabular}
\label{tab:gans_accuracy}
\end{table}

\begin{figure}
    \centering
    \begin{subfigure}{0.30\textwidth}
        \includegraphics[width=\textwidth]{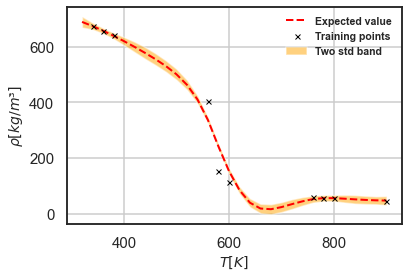}
    \end{subfigure}
    \begin{subfigure}{0.30\textwidth}
        \includegraphics[width=\textwidth]{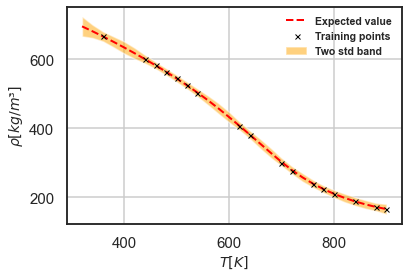}
    \end{subfigure}
    \begin{subfigure}{0.30\textwidth}
        \includegraphics[width=\textwidth]{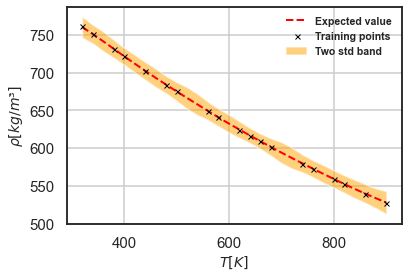}
    \end{subfigure}
    
    \begin{subfigure}{0.30\textwidth}
        \includegraphics[width=\textwidth]{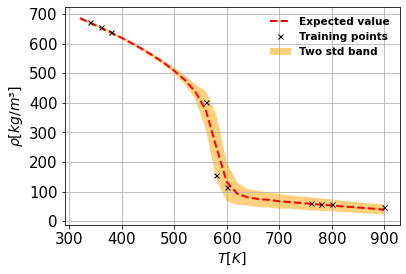}
        \caption{3 MPa}
    \end{subfigure}
    \begin{subfigure}{0.30\textwidth}
        \includegraphics[width=\textwidth]{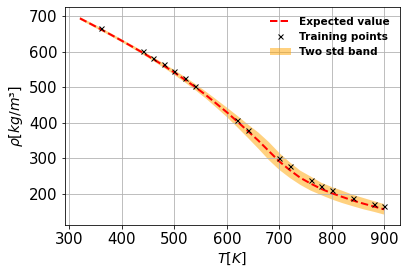}
        \caption{10 MPa}
    \end{subfigure}
    \begin{subfigure}{0.30\textwidth}
        \includegraphics[width=\textwidth]{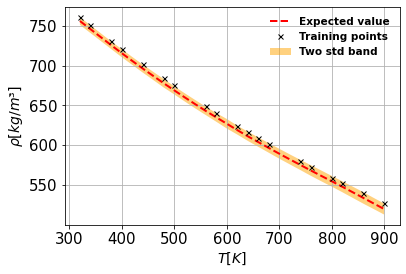}
        \caption{100 MPa}
    \end{subfigure}
    \caption{n-Octane predictions with the GP (top) and probabilistic conditional generative models (bottom) at the pressures 3, 10, and 100 MPa.}
    \label{fig:octane_50_gp_gans}
\end{figure}

{As a further illustration of the performance of such approaches to predict the density, we plot its values for  n-octane, n-dodecane, and n-hexadecane densities with respect to temperature for the ML models trained with 50\% of the dataset, since this training scenario returns the best relation between accuracy and computational cost. Figure \ref{fig:octane_50_gp_gans} shows the n-octane density predictions  at the pressures equal to 3, 10, and 100 MPa. We can observe that at 3MPa the GP model fails to deliver good results around the transcritical region, while the  generative model provides robust predictions with uncertainties bounds that capture the data. The predictive uncertainty of the proposed approaches reflects limited data for training the models, the epistemic uncertainty. We can also note that both models perform well at 10 and 100 MPa, wherein the density dependency on the temperature has a smooth behavior.}

\begin{figure}
    \centering
    \begin{subfigure}{0.30\textwidth}
        \includegraphics[width=\textwidth]{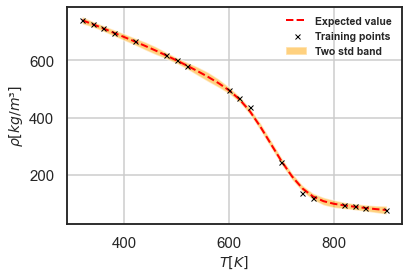}
    \end{subfigure}
    \begin{subfigure}{0.30\textwidth}
        \includegraphics[width=\textwidth]{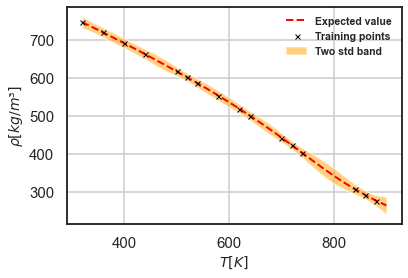}
    \end{subfigure}
    \begin{subfigure}{0.30\textwidth}
        \includegraphics[width=\textwidth]{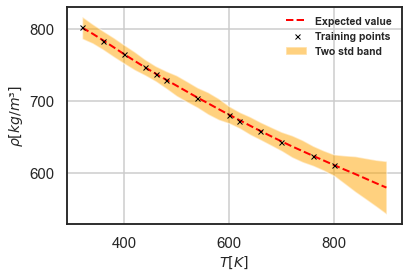}
    \end{subfigure}
    
    \begin{subfigure}{0.30\textwidth}
        \includegraphics[width=\textwidth]{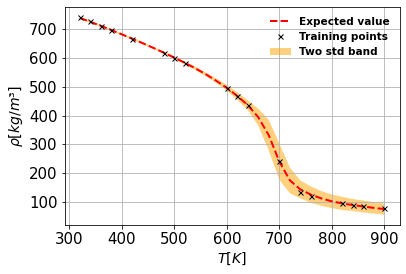}
        \caption{3 MPa}
    \end{subfigure}
    \begin{subfigure}{0.30\textwidth}
        \includegraphics[width=\textwidth]{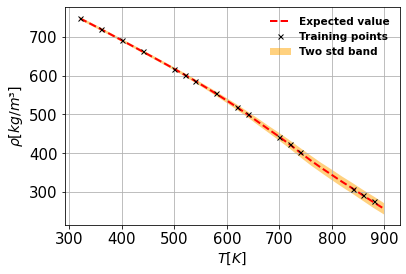}
        \caption{10 MPa}
    \end{subfigure}
    \begin{subfigure}{0.30\textwidth}
        \includegraphics[width=\textwidth]{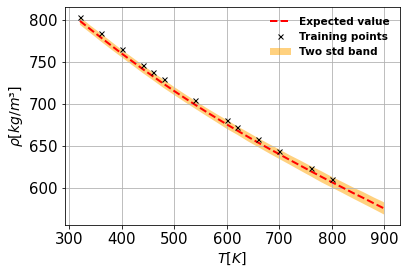}
        \caption{100 MPa}
    \end{subfigure}
    \caption{n-Dodecane predictions with the GP machine learning model (top) and conditional generative machine learning model (bottom) at the pressures 3, 10, and 100 MPa.}
    \label{fig:dodecane_50_gp_gans}
\end{figure}

\begin{figure}
    \centering
    \begin{subfigure}{0.30\textwidth}
        \includegraphics[width=\textwidth]{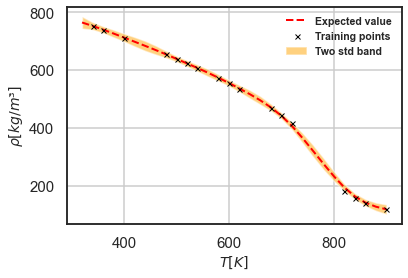}
    \end{subfigure}
    \begin{subfigure}{0.30\textwidth}
        \includegraphics[width=\textwidth]{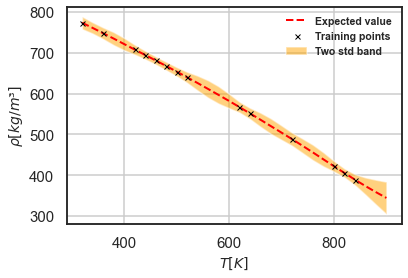}
    \end{subfigure}
    \begin{subfigure}{0.30\textwidth}
        \includegraphics[width=\textwidth]{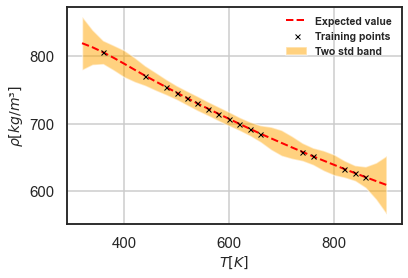}
    \end{subfigure}
    
    \begin{subfigure}{0.30\textwidth}
        \includegraphics[width=\textwidth]{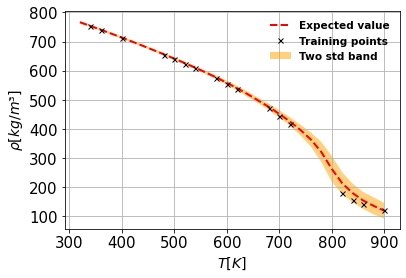}
        \caption{3 MPa}
    \end{subfigure}
    \begin{subfigure}{0.30\textwidth}
        \includegraphics[width=\textwidth]{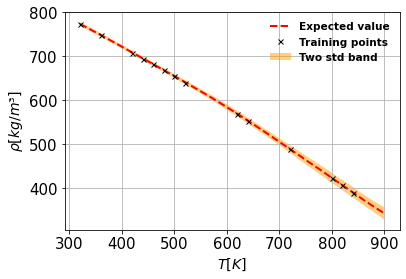}
        \caption{10 MPa}
    \end{subfigure}
    \begin{subfigure}{0.30\textwidth}
        \includegraphics[width=\textwidth]{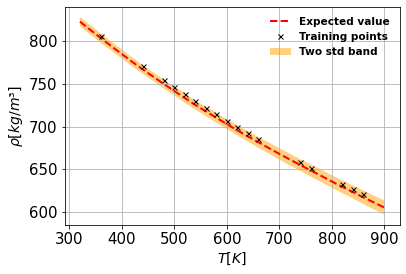}
        \caption{100 MPa}
    \end{subfigure}
    \caption{n-Hexadecane predictions with the GP machine learning model (top) and conditional generative machine learning model (bottom) at the pressures 3, 10, and 100 MPa.}
    \label{fig:hexadecane_50_gp_gans}
\end{figure}

Also, the n-dodecane and n-hexadecane densities are depicted along with temperature in Figs. \ref{fig:dodecane_50_gp_gans} and \ref{fig:hexadecane_50_gp_gans}. We observe that the ML models return robust predictions at three different pressures. Besides, it is noted that the GP model returns larger uncertainty bounds at high pressures, specifically at density points not used in the training process.

\begin{figure}
    \centering
    \begin{subfigure}{0.30\textwidth}
        \includegraphics[width=\textwidth]{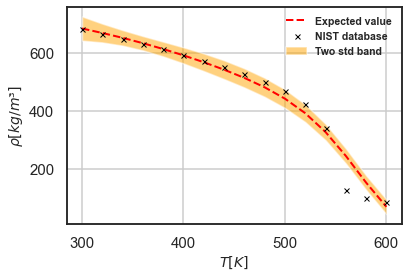}
    \end{subfigure}
    \begin{subfigure}{0.30\textwidth}
        \includegraphics[width=\textwidth]{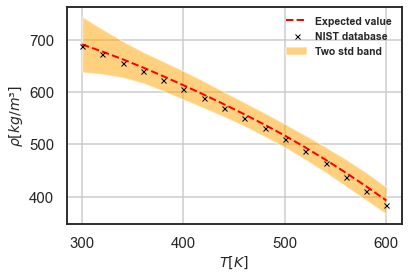}
    \end{subfigure}
    \begin{subfigure}{0.30\textwidth}
        \includegraphics[width=\textwidth]{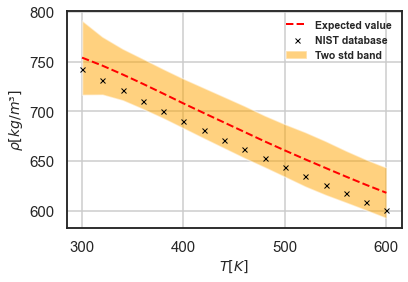}
    \end{subfigure}
    
    \begin{subfigure}{0.30\textwidth}
        \includegraphics[width=\textwidth]{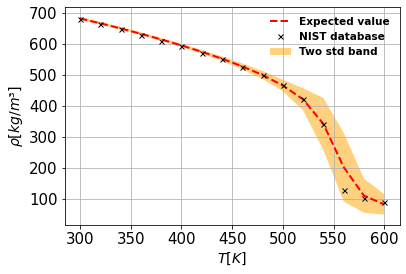}
        \caption{3 MPa}
    \end{subfigure}
    \begin{subfigure}{0.30\textwidth}
        \includegraphics[width=\textwidth]{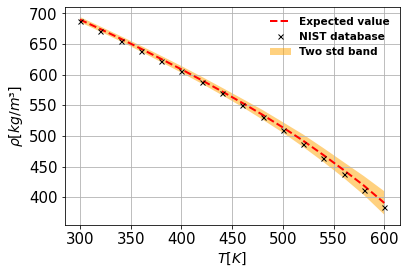}
        \caption{10 MPa}
    \end{subfigure}
    \begin{subfigure}{0.30\textwidth}
        \includegraphics[width=\textwidth]{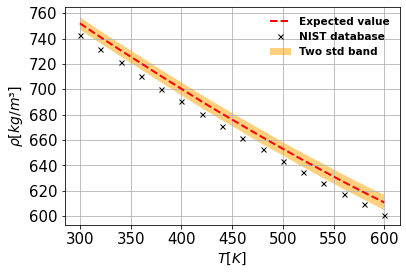}
        \caption{100 MPa}
    \end{subfigure}
    \caption{n-Heptane predictions with the GP machine learning model (top) and conditional generative machine learning model (bottom) at the pressures 3, 10, and 100 MPa.}
    \label{fig:heptane_50_gp_gans}
\end{figure}

We also test how the proposed ML technology perform in an extrapolation scenario. We test them for the n-heptane, a fuel not used for building the models. In order to do that, instead of employing data provided by ML computations, we use an experimental database furnished by the National Institute of Standards and Technology (NIST). Figure \ref{fig:heptane_50_gp_gans} shows that at 3 MPa and liquid condition the ML model returns good predictions of the n-heptane density behavior, with small uncertainties. However, at supercritical conditions ($T_c = 540.13K$), the GP model returns density predictions far from satisfactory. Also, we note that the generative model has uncertainty bounds able to capture the thermophysical property. The $L_2$ mean relative error between the NIST dataset and the expected values predicted by the GP and conditional generative models are $7.1697\times 10^{-2}$ and $2.0838\times 10^{-2}$, respectively. We can also note that at higher pressure where the density behavior is smooth, the models present better predictions, with the GP model showing larger uncertainties bounds and the generative model returns smaller uncertainty bounds. Moreover, the $L_2$ mean relative errors of the GP model at 10 and 100 MPa are respectively $1.8152\times 10^{-4}$ and $6.3072\times 10^{-4}$, and for the conditional generative model the $L_2$ mean relative errors at the same pressures are $8.4484\times 10^{-5}$ and $2.0322\times 10^{-4}$.

{Furthermore, we use a coefficient of variation to measure the degree of uncertainty of the density predictions. It is defined as the ratio between the standard deviation $\sigma_{\rho}$ and the mean $\mu_{\rho}$ of the prediction}

\begin{equation}
    \text{cv}({p,T}) = \frac{\sigma_{\rho}({p,T})}{\mu_{\rho}({p,T})}
\end{equation}

\begin{figure}
     \centering
     \begin{subfigure}{0.45\textwidth}
         \includegraphics[width=\textwidth]{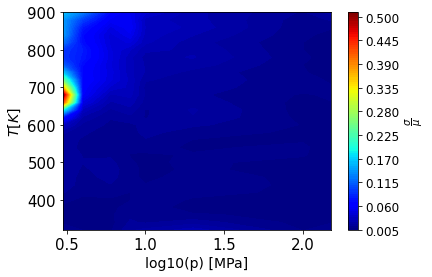}
         \caption{GP model}
     \end{subfigure}
     \begin{subfigure}{0.45\textwidth}
         \includegraphics[width=\textwidth]{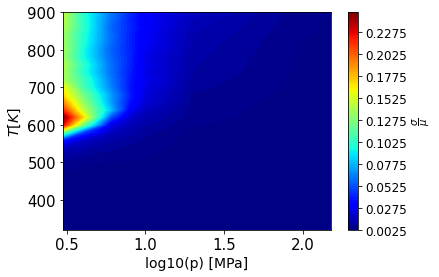}
         \caption{Conditional generative model}
     \end{subfigure}
     \caption{n-Octane density variability for a range of temperatures and pressures.}
     \label{fig:octane_rmap_diagram}
\end{figure}

{Figure \ref{fig:octane_rmap_diagram} gives an overall picture by displaying a mapping between the operating conditions and the uncertainty on n-octane density predictions. We present an explicit quantification of the epistemic uncertainty resulting from the lack of data, which helps to understand limits of the ML models. More specifically, to make more accessible the visualization of the results, we plot this mapping for $\text{log}_{10}p \in [0.5,2.5]$ MPa and $T \in [320, 900]$ K with regular intervals of 20K, allowing us to make explicit the strong dependence of the epistemic uncertainties regarding different regions of  operating conditions. A critical aspect to be remarked is the higher values of $\text{cv}$ in particular regions of the operating conditions space, especially at transcritical conditions displaying  higher gradients of the  property. We can note that the GP model returns a degree of uncertainty slightly large in this region.  That can be mitigated by providing more training data for this specific region. Also, it is noted that variability of density provided by the conditional generative model is less pronounced at liquid regions and for high-pressure supercritical regions, which is due to the smooth density behavior resulting in a low degree of uncertainty in the predictions at these regions.}


\subsection{Data-fusion Machine Learning models}

\begin{figure}
    \centering
        \includegraphics[width=0.40\textwidth]{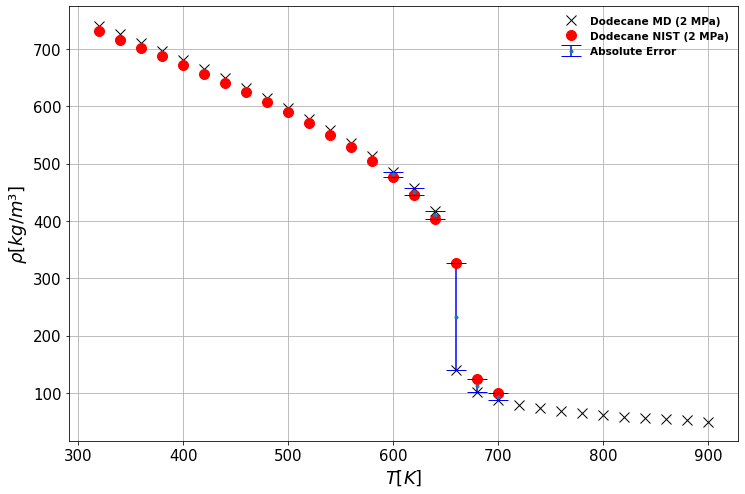}
    \caption{Comparison between n-dodecane density predictions along with temperature at 2 MPa between MD simulations against NIST dataset.}
    \label{fig:density_dodecane_error}
\end{figure}

\begin{figure}
     \centering
     \begin{subfigure}{0.35\textwidth}
         \includegraphics[width=\textwidth]{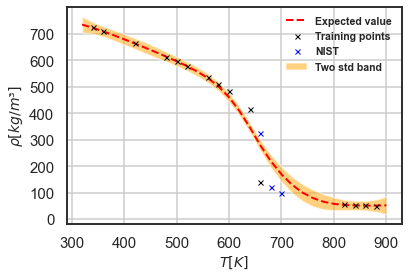}
         \caption{GP model}
     \end{subfigure}
     \begin{subfigure}{0.35\textwidth}
         \includegraphics[width=\textwidth]{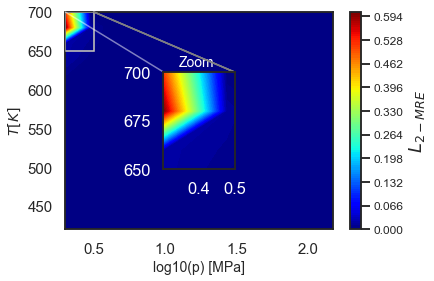}
         \caption{$L_2$ relative error}
     \end{subfigure}
     \caption{n-Dodecane density predictions GP model: (a) n-Dodecane density along with temperature at 2 MPa. (b) $L_2$ error between the expected value predicted by the ML model against NIST.}
     \label{fig:dodecane_error_diagram_GP}
\end{figure}

\begin{figure}
     \centering
     \begin{subfigure}{0.35\textwidth}
         \includegraphics[width=\textwidth]{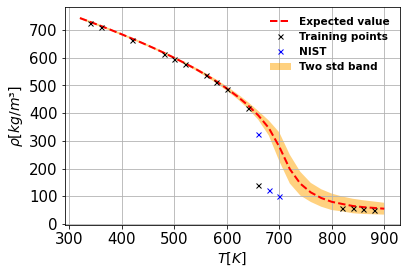}
         \caption{Conditional generative model}
     \end{subfigure}
     \begin{subfigure}{0.35\textwidth}
         \includegraphics[width=\textwidth]{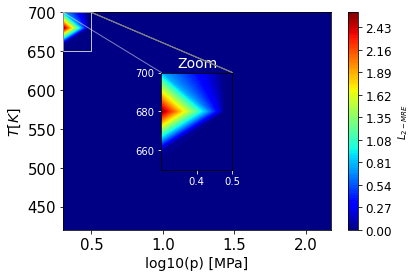}
         \caption{$L_2$ relative error}
     \end{subfigure}
     \caption{n-Dodecane density predictions conditional generative model: (a) n-Dodecane density along with temperature at 2 MPa. (b) $L_2$ error between the expected value predicted by the ML model against NIST.}
     \label{fig:dodecane_error_diagram_GANs}
\end{figure}

{Although MD simulation is considered  to be a robust tool to predict thermodynamic properties, it returns incorrect values at critical points/transcritical regions. It was shown \cite{CHEN201948} that the transport properties predictions of diesel surrogate fuels are far from satisfactory near such  critical points. That is also the case with n-dodecane  in that study. Density predictions with MD simulations and NIST data at transcritical regions present considerable discrepancies, as shown in Fig. \ref{fig:density_dodecane_error}. More specifically, in operating conditions near the critical point of n-dodecane,  critical pressure ($P_c=1.8170$ MPa) and critical temperature ($T_c = 658.1$ K), our ML models based on the MD data fail to accurately predict the density. Figures \ref{fig:dodecane_error_diagram_GP} (a) and \ref{fig:dodecane_error_diagram_GANs} (a) show the density predictions at 2 MPa for GP and conditional generative model, respectively. Moreover, Figures \ref{fig:dodecane_error_diagram_GP} (b) and \ref{fig:dodecane_error_diagram_GANs} (b) also show that the main discrepancies between the expected values of ML models against the NIST database are into the transcritical regions.}

Aiming at improving the predictability of our ML models at transcritical regions, we adopt two  strategies, exploring the fusion of MD simulations with experimental data. The aim here is not to compare these different strategies but to evaluate their potential. Both are formulated with the same idea, promoting the fusion of data from MD simulations and experiments datasets. In the first one, we propose a data-fusion strategy in which density points of the transcritical region provided by the NIST database are simply concatenated into the training dataset. The second differs as we propose a multi-fidelity arrangement of the data.  A detailed description of both strategies is given further ahead.

\begin{figure}
    \centering
    \begin{subfigure}{0.35\textwidth}
        \includegraphics[width=\textwidth]{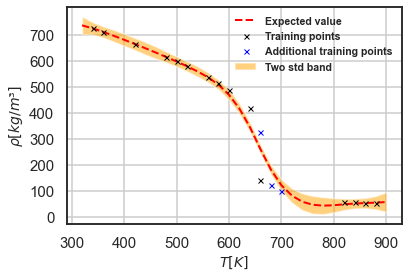}
    \end{subfigure}
    \begin{subfigure}{0.35\textwidth}
        \includegraphics[width=\textwidth]{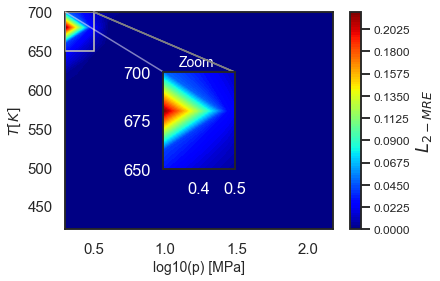}
    \end{subfigure}
    \begin{subfigure}{0.35\textwidth}
        \includegraphics[width=\textwidth]{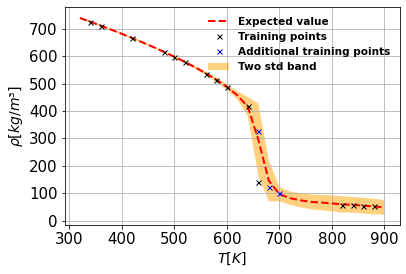}
        \caption{2 MPa}
    \end{subfigure}
    \begin{subfigure}{0.35\textwidth}
        \includegraphics[width=\textwidth]{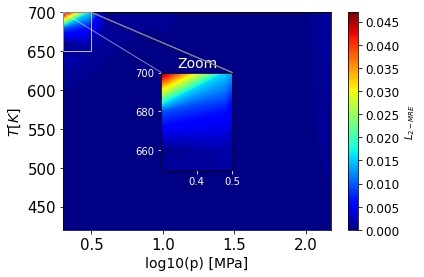}
        \caption{L2 relative error}
    \end{subfigure}
    \caption{n-Dodecane density predictions with the GP model (top) and conditional generative model (bottom) using the data-fusion approach with three density points from the NIST database.}
    \label{fig:dodecane_MD_3point_NIST}
\end{figure}

In the data-fusion approach, we add three density values from NIST to the original training dataset, as depicts in Fig \ref{fig:dodecane_MD_3point_NIST} (a). Note that the fusion improves considerably the predictions of the conditional generative model with relative errors lower than 5\%, while the GP model still returns relative errors not satisfactory. Further details about this data-fusion approach can be found in the Appendix \ref{App:data_fusion_studies}. 

As discussed above, generating reliable data with MD simulations to be used in supervised learning might require a great computational effort. To tackle such a drawback, numerical formulations combining models displaying different levels of fidelity are frequently employed. Those multi-fidelity simulators employ, for instance, coarse grid discretizations, models based on simplified physics, or simplified iterative methods. Here, again we merge experimental data with MD simulations, restricting our approaches to two levels of fidelity.

In this new context, we propose extensions of the previous introduced ML models. We start by obtaining high-fidelity $\{\mathbf{x}_H,\gamma_H\}$ and low-fidelity $\{\mathbf{x}_L,\gamma_L\}$ input-output samples. Typically,  the number of samples in the first case tends to be much smaller due to the related costs. We assign the high-fidelity score to the experimental data, according to the considerations above about the potential inaccuracy of the MD obtained computed properties for transcritical regions.

We start with our first, in this multi-fidelity context,  ML model   approximating the conditional probability $p(\gamma_H |\mathbf{x}_H,\gamma_L,z)$, using the generative model $\gamma_H=f_{\phi}(\mathbf{x}_H,\gamma_L,z)$, $z\sim p(z)$. In another words, the ML model is supposed to capture the correlation between the two level of fidelity data. Once this is achieved, we have a predictive model computing outputs for a new  point
$\mathbf{x}^*$: $\mathbf{y_H}^* = f_{\phi}(\mathbf{x}^*,f_L(\mathbf{x}^*),z)$. At this point, it is worth remarking that  one of the inputs  is the output of the low-fidelity model, leading to a recursive scheme to obtain the predictions of the multifidelity model. In fact, here the considered low fidelity data is produced with expensive MD simulations. Therefore, in order to achieve a feasible scheme, we need to build an auxiliary, cheap to compute and accurate, proxy for the low fidelity model using the available data.

As a second approach, the one based on GPs, we employ the nonlinear autoregressive multi-fidelity GP  (NARGP) regression model \cite{RAISSI0751}. The main idea of the NARGP model is to extend  GP modeling to capture nonlinear correlations from data generated by sources of different fidelity \cite{LeGratiet_2014,OHagan_GP}. It  enables the construction of  probabilistic models prone to encapsulate uncertainties, built upon the recursive relation  $y_H = g (x_H, f_L(x_H))$ involving low and high fidelity data, in which $f_L$ is a GP model for the former. Moreover, we put a GP prior on $g$. After the training, we obtain the predictive model, which turns to be also a GP, $y_H = g (x^*, f_L(x^*))$.

To assess the above multi-fidelity ML approaches, we use an illustrative example involving data  from "low-fidelity" MD simulations and "high-fidelity" NIST experimental values. For both approaches, the training dataset consists of 7 density values of n-dodecane $\rho_H(p,T_H)$ and $\rho_L(p,T_L)$, at the pressure of $2$ MPa and a set of temperatures given by $T_H=T_L =\{320,  440, 500, 620, 660, 680, 700\}$ K. Note that we prioritize points located in the transcritical part, since this region presents larger discrepancies between the values predicted by MD simulations and the NIST database.

The conditional generative model is constructed using fully connected feed-forward architecture for the encoder and generator networks with 4 hidden layers and 100 neurons per layer, while the discriminator architecture has 2 hidden layers with 100 neurons per layer. All activation uses a hyperbolic tangent non-linearity. The models are trained for 20,000 stochastic gradient descent steps using the Adam optimizer \cite{kingma2017adam} with a learning rate of $10^{-4}$, while fixing a one-to-five ratio for the discriminator versus generator updates. Furthermore, we have fixed the entropy regularization parameter to $\lambda=1.5$, and we also employed a one-dimensional latent space  with a standard normal prior, $p(z) \sim \mathcal{N}(0, 1)$.

We train the NARGP model via maximizing the marginal log-likelihood using the Mattern 3/2 kernel function. The gradient descend optimizer L-BFGS is used considering randomized restarts to ensure convergence to a global optimum. Once the high-fidelity recursive GP is trained, we can compute the predictive posterior mean and variance at a given untested point $\mathbf{x}^*$ by sampling the probabilistic predictive model. 

\begin{figure}
     \centering
     \begin{subfigure}{0.45\textwidth}
        \includegraphics[width=\textwidth]{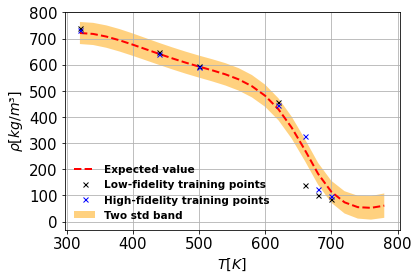}
        \caption{NARGP model}
     \end{subfigure}
     \begin{subfigure}{0.45\textwidth}
        \includegraphics[width=\textwidth]{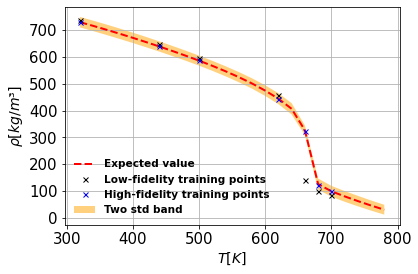}
        \caption{Conditional generative model}
     \end{subfigure}
     \caption{Multi-fidelity modeling of n-dodecane diesel surrogate fuel density.}
     \label{fig:MF_NARGP_GANS}
\end{figure}

The main results are  summarized in Fig \ref{fig:MF_NARGP_GANS}. More specifically, the results indicate that the NARGP model was able to satisfactorily reconstruct the high-fidelity data. To make this comparison quantitative, we compute the mean $L_2$ relative error between the expected values predicted by the generative model and the NIST data. It shows predictions with accuracy of $L_{2-MRE}=1.4524 \times 10^{-2}$.  Moreover, it returns good uncertainty bounds able to capture the high-fidelity response at the transcritical region. Also, we note a perfect agreement between the expected value provided by the probabilistic conditional generative model and the high-fidelity data, resulting in an accuracy of $L_{2-MRE}=4.4782 \times 10^{-5}$. Finally, we observe that the multi-fidelity model returns small uncertainty bounds despite the small amount of data employed in the training process. 

\section{Conclusions}

In this work, we propose a computational methodology based on the use of ML with Molecular Dynamics simulations to compute physicochemical properties of single compound fuels at engine-relevant conditions. The ML models have revealed to be a powerful tool to accurately predict the fuel properties from the chemical composition for a wide range of alkanes. Moreover, this study explores the versatility of the ML models to handle data from different sources, which can then be integrated efficiently in the context of UQ workflows with many-query tasks.

We place our contribution in the emerging area of physics-aware ML, where the final model, in many different ways, blends two main components: availability of experimental data and/or often expensive computational models relying on first principles and phenomenological closure equations, and deep learning data-driven models. Such combination allows describing physicochemical properties over a wide range of flow conditions at relatively low cost, and also offers a broad spectrum of opportunities to enhance CFD codes.

This study has shown a successful prediction of physical quantities, in this case density, that can also be extended to more complex fuel molecules or multicomponent mixtures like dimethyl ethers or oxymethylene dimethyl ethers. The generation of  reliable physicochemical properties of renewable fuels is an important step forward towards the generation of digital tools that can assist on the decarbonization by the use of renewable fuels.

\bibliographystyle{unsrtnat}


\begin{appendices}
\section{Data-fusion studies}
\label{App:data_fusion_studies}

\begin{figure}[h]
    \centering
    \begin{subfigure}{0.35\textwidth}
        \includegraphics[width=\textwidth]{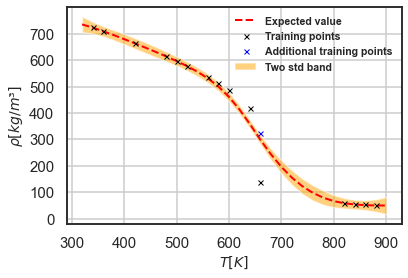}
    \end{subfigure}
    \begin{subfigure}{0.35\textwidth}
        \includegraphics[width=\textwidth]{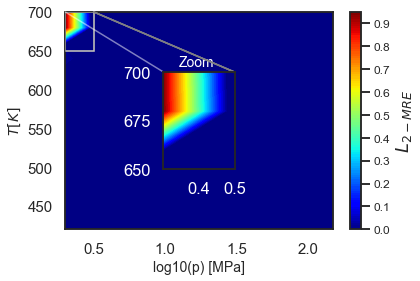}
    \end{subfigure}
    \begin{subfigure}{0.35\textwidth}
        \includegraphics[width=\textwidth]{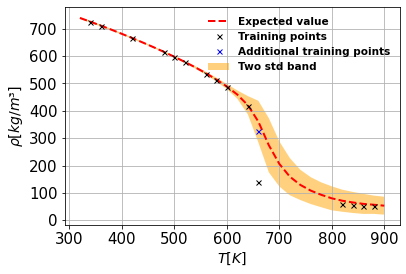}
        \caption{2 MPa}
    \end{subfigure}
    \begin{subfigure}{0.35\textwidth}
        \includegraphics[width=\textwidth]{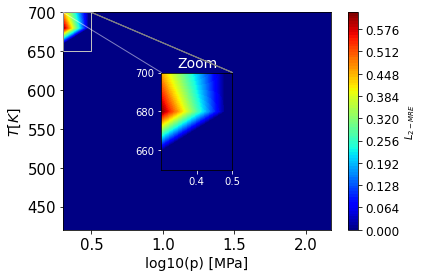}
        \caption{$L_2$ relative error}
    \end{subfigure}
    \caption{n-Dodecane density predictions with the GP model (top) and conditional generative model (bottom) using the data-fusion-fidelity approach with one density point from the NIST database.}
    \label{fig:dodecane_MD_1point_NIST}
\end{figure}

\begin{figure}
    \centering    
    \begin{subfigure}{0.35\textwidth}
        \includegraphics[width=\textwidth]{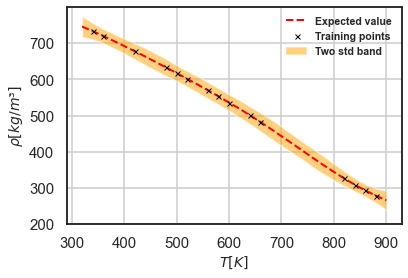}
    \end{subfigure}
    \begin{subfigure}{0.35\textwidth}
        \includegraphics[width=\textwidth]{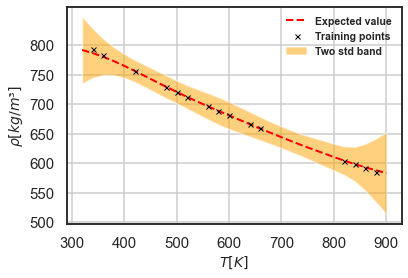}
    \end{subfigure}
    \begin{subfigure}{0.35\textwidth}
        \includegraphics[width=\textwidth]{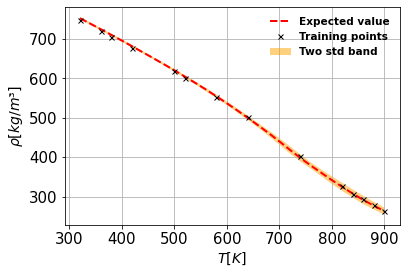}
        \caption{10 MPa}
    \end{subfigure}
    \begin{subfigure}{0.35\textwidth}
        \includegraphics[width=\textwidth]{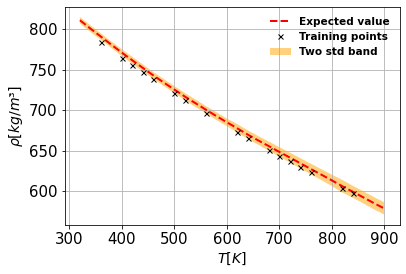}
        \caption{100 MPa}
    \end{subfigure}
    \caption{n-Dodecane density predictions with the GP model (top) and conditional generative model (bottom) using the data-fusion-fidelity approach with one density point from the NIST database at the pressures 10 and 100 MPa.}
    \label{fig:dodecane_MD_1point_NIST_10_100}
\end{figure}

In order to enhance the predictability of the ML models at transcritical regions, here we propose a data-fusion approach. Specifically, we concatenate density points of the transcritical region provided by the NIST database into the training dataset. The aim here is to improve the density predictability of our ML models, by supplying reliable information about this state variable in the specific region where MD data is scarce. Following this purpose, the first attempt is to add one density point from the NIST database. Here, we concatenate the n-dodecane density at pressure 2 MPa and temperature 660 K to the training data. By adding this point to the training set, it is verified that the ML models can recover the density at 660 K, as shown in Fig. \ref{fig:dodecane_MD_1point_NIST}. However, the $L_2$ relative errors between the expected values predicted by ML models and the NIST data are still considerable in transcritical regions. Also, we can note that the conditional generative model has larger uncertainty bounds at the transcritical region trying to recover density behavior due to the lack of data in this region. Furthermore, Figure \ref{fig:dodecane_MD_1point_NIST_10_100} shows that adding density points from NIST into the training data does not change the degree of uncertainty at other operating conditions.

\begin{figure}
    \centering
    \begin{subfigure}{0.35\textwidth}
        \includegraphics[width=\textwidth]{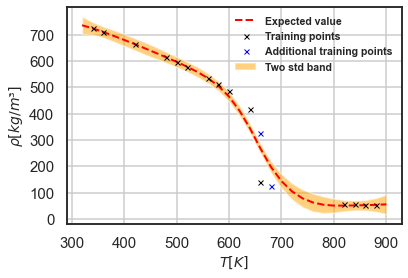}
    \end{subfigure}
    \begin{subfigure}{0.35\textwidth}
        \includegraphics[width=\textwidth]{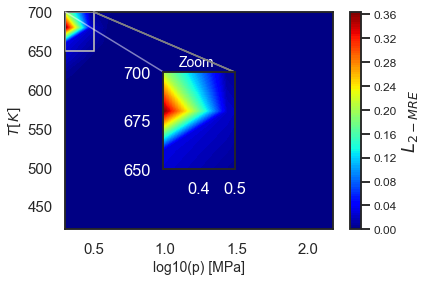}
        
    \end{subfigure}
    \begin{subfigure}{0.35\textwidth}
        \includegraphics[width=\textwidth]{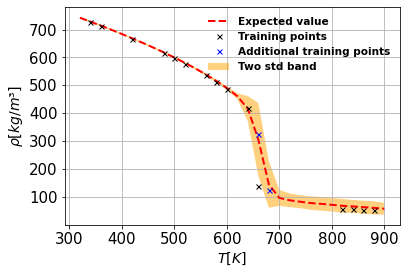}
        \caption{2 MPa}
    \end{subfigure}
    \begin{subfigure}{0.35\textwidth}
        \includegraphics[width=\textwidth]{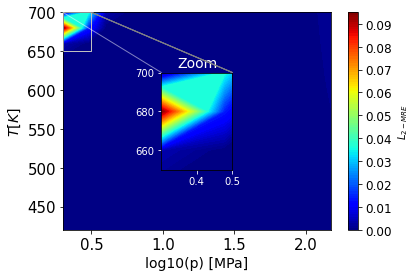}
        \caption{L2 relative error}
    \end{subfigure}
    \caption{n-Dodecane density predictions with the GP model (top) and conditional generative model (bottom) using the data-fusion-fidelity approach with two density points from the NIST database.}
    \label{fig:dodecane_MD_2point_NIST}
\end{figure}

As a further attempt to enhance the density predictions at the transcritical region, we now concatenate one more density point from the NIST database. More specifically, in addition to concatenating the n-dodecane density at pressure 2 MPa and temperature 660 K to the training data, we also add the n-dodecane density at 680 K. Figure \ref{fig:dodecane_MD_2point_NIST} shows that adding two density points from NIST data in the transcritical region slightly improves the predictions of the GP model, while the relative error remains considerable. However, we can verify that the generative model returns satisfactory predictions with $L_2$ relative error lower than 10\% in the transcritical region. This shows the capability of the conditional generative model to enhance the predictability of the density when some pieces of information about the correct behavior of the transport property are given to the model.

Finally, to further increase the predictability of our ML models, a third attempt is proposed based on adding three density points from NIST to the training data, those being the n-dodecane densities at 2 MPa and temperatures 660, 680, and 700K. Figure \ref{fig:dodecane_MD_3point_NIST} depicts that in this training scenario the density predictions of the GP model have some improvements, but the $L_2$ relative error is still considerable. Furthermore, we can verify that the conditional generative model returns accurate predictions, with relative errors lower than 5\% in transcritical regions. Finally, we note that the generative model has uncertainty bounds able to recover the density predictions near the critical point.
\end{appendices}

\end{document}